\documentclass[final]{cvpr}

\usepackage{times}
\usepackage{epsfig}
\usepackage{graphicx}
\usepackage{amsmath}
\usepackage{amssymb}
\usepackage[inline,shortlabels]{enumitem}
\usepackage{url}

  \usepackage{float}
  \usepackage{comment}
  \usepackage{empheq}

  \usepackage{algorithm,algorithmic}
  \usepackage{subfig}
  \usepackage{multirow}
  
\usepackage{makecell}
 \usepackage{xcolor}
 \usepackage{subfig}

\usepackage[pagebackref=true,breaklinks=true,colorlinks,bookmarks=false]{hyperref}
\usepackage{cleveref}

\def\cvprPaperID{40} %
\def\confYear{CVPR 2021}
\newcommand*{\affaddr}[1]{#1} %
\newcommand*{\affmark}[1][*]{\textsuperscript{#1}}
\newcommand*{\email}[1]{\texttt{#1}}

\begin{document}

\title{All You Can Embed: Natural Language based Vehicle Retrieval with Spatio-Temporal Transformers }

\author{%
Carmelo Scribano\affmark[\dag]\affmark[*], Davide Sapienza\affmark[\dag]\affmark[*], Giorgia Franchini\affmark[\dag]\affmark[\ddag], Micaela Verucchi\affmark[\dag], and Marko Bertogna\affmark[\dag]\\
\affaddr{\affmark[\dag]University of Modena and Reggio Emilia}
\affaddr{\affmark[\ddag]University of Ferrara}
\affaddr{\affmark[*]University of Parma}\\
\email{\{name.surname\}@unimore.it}\\
}

\maketitle

\begin{abstract}
Combining Natural Language with Vision represents a unique and interesting challenge in the domain of Artificial Intelligence.
The AI City Challenge Track 5 for Natural Language-Based Vehicle Retrieval focuses on the problem of combining visual and textual information, applied to a smart-city use case. 
In this paper, we present All You Can Embed (AYCE), a modular solution to correlate single-vehicle tracking sequences with natural language. The main building blocks of the proposed architecture are (i) BERT to provide an embedding of the textual descriptions, (ii) a convolutional backbone along with a Transformer model to embed the visual information. 
For the training of the retrieval model, a variation of the Triplet Margin Loss is proposed to learn a distance measure between the visual and language embeddings.
The code is publicly available at \url{https://github.com/cscribano/AYCE_2021}.

\end{abstract}
\section{Introduction} 
\label{sec:introduction}

Vision and Natural Language (NL) understanding is a relevant and challenging problem in computer vision. Recently, researchers have been focusing on binding video and NL to deal with tasks such as video captioning, action segmentation, video question answering, or text-based video retrieval.  
In this paper, we focus on a smart-city-specific variation of the latter problem, i.e. the NL-based vehicle retrieval, introduced by Feng,  Ablavsky, and Sclaroff~\cite{feng2021cityflow} for the AI City Challenge Track 5\footnote{https://www.aicitychallenge.org/2021-challenge-tracks/}.
The task under analysis combines the need for spatio-temporal coherence with the need to correlate video and NL, and is therefore non-trivial. 
However, solving the problem would allow retrieving traffic patterns or specific vehicle-related events just by typing a sentence in a camera system, a convenient and useful feature for people involved in urban planning, traffic engineering, or law enforcement. 

\begin{figure}
    \subfloat[``A white pick up truck drives down the street and passes another truck.'', ``A white pickup truck going straight down the street with cars parking on the side.'', ``A white pickup runs on the street.'']{\includegraphics[clip, trim=0.5cm 17cm 0.5cm 3cm, width=0.5\textwidth]{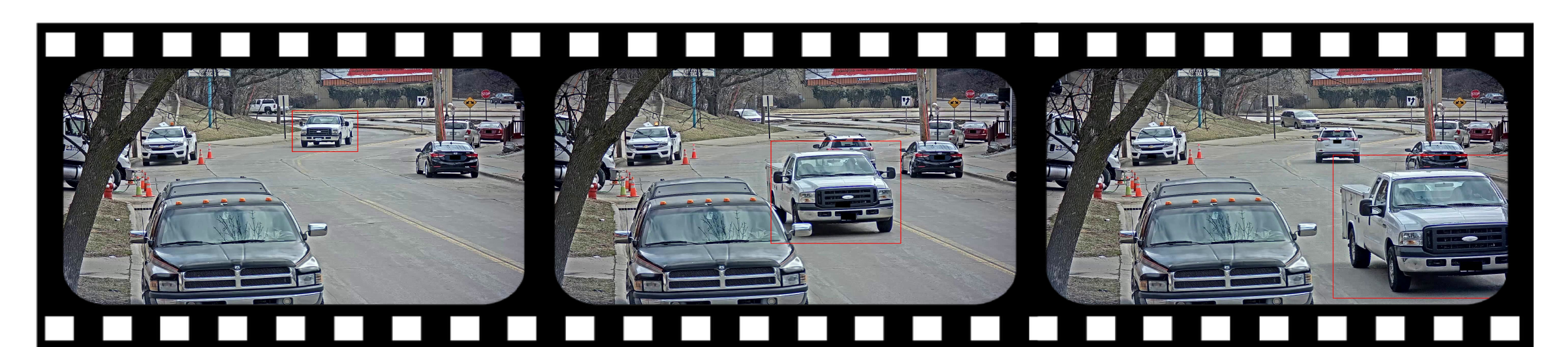}}\\

    \caption{A sample from the CityFlow-NL dataset.}
    \label{fig:dataset}
\end{figure}

Inspired by other works that combine NL with images~\cite{herdade2019image} or video~\cite{sun2019videobert, sun2019videobert, zhu2020actbert, miech2020end, luo2020univilm, li2020hero, lei2021less}, we developed All You Can Embed (AYCE), an NL-based vehicle retrieval system for a target object. 
Our approach builds on the popular BERT (Bidirectional Encoder Representations from Transformers)
model~\cite{devlin2018bert} to handle NL Processing (NLP). 
Besides, we use two other Transformers to encode the visual features: a first one to embed the spatial information of a single frame, and a second one to embed the temporal information throughout all the frames of the video sequence.
The architecture has been tested on the CityFlow-NL~\cite{feng2021cityflow} dataset; \Cref{fig:dataset}
 depicts a sample out of its 2.5k sequences.
 
Our contributions are summarized as follows:
\begin{enumerate}
    \item We introduce a stacked spatio-temporal Transformer Encoder to embed the frame-level spatial information first, and the video-level temporal one later.
    \item We adopt a custom loss, obtained from a variation of the Triplet Margin Loss (TML)~\cite{schultz2004learning}, to correlate the visual and NL information. The objective is to keep corresponding single-vehicle tracking sequence and NL embeddings as close as possible.
    \item In conclusion, we propose a novel architecture to combine visual and NL information for the Vehicle Retrieval task, given three sentences (or sentence-triplet). 

\end{enumerate}

\Cref{sec:relatedworks} gives an overview of the state-of-the-art works we took inspiration from, while \Cref{sec:dataanalysis} reports some insights of the specific task and dataset. \Cref{sec:method} describes the main contribution of this paper, giving all the details of the AYCE architecture and, finally, a selection of significant experiments can be found in \Cref{sec:experiments}.

\section{Related Works} 
\label{sec:relatedworks}

In the literature, the standard approach to extract meaningful information from visual and textual sources is to generate proper embeddings. Over the years, several methods have been proposed to obtain embeddings from images or videos, from NL, or their combination.

\subsection{Language Modeling}
Regarding NL, pre-trained sentence encoders such as ELMo~\cite{peters2018deep} and BERT~\cite{devlin2018bert} have rapidly improved the state of the art on many NLP tasks, e.g. question answering or natural language inference, dominating the previous solutions based on context-independent word embeddings such as word2vec~\cite{mikolov2013efficient} and Glove~\cite{pennington2014glove}. The novelty introduced by BERT is the use of the Transformer architecture instead of LSTM (used for example in ELMo). 
Since its introduction, there has been continuous advancement in language model pre-trainings.

\subsection{Visual features embedding}

To extract visual information, a common strategy is to apply Convolutional Neural Networks (CNNs). CNNs have been proposed to solve classic computer vision tasks such as image classification or object detection. To do so, those models learn an inner representation of the visual source.

Among the various existing CNNs, it is worth mentioning R-CNN~\cite{girshick2014rich}, Fast R-CNN~\cite{girshick2015fast}, and Faster R-CNN~\cite{ren2015faster}, a family of regional convolution networks. 
The name of these models is due to the presence of an explicit region proposal network that proposes regions from convolution features, picking the ones that are likely to carry the most meaningful visual information of the image. Such latent representation is then given to a classifier for the second step of object detection, reason why R-CNNs are known as two-staged object detectors. 

Given their success in extracting valuable knowledge, classifiers and object detectors have also been applied to multi-modal tasks such as image or video captioning or video understanding. 

For example, Herdade et al.~\cite{herdade2019image} make use of Faster R-CNN to encode the visual information of an image and transform objects into words. As a further step, the authors introduce the Object Relation Transformer, an encoder-decoder architecture designed specifically for image captioning, that incorporates information about the spatial relationships between input detected objects through geometric attention. 
Indeed, even though Transformers have been originally proposed for NLP tasks, they have also been successfully applied, alone or in conjunction with a convolutional part, in a wide variety of vision tasks such as classification ~\cite{dosovitskiy2020image}, object detection ~\cite{carion2020end}, object tracking ~\cite{sun2020transtrack}, and others.

\subsection{Joint video-text embedding}
\begin{figure}[!ht]
    \subfloat[]{\includegraphics[clip, trim=1cm 9cm 17cm 0cm, width=0.16\textwidth]{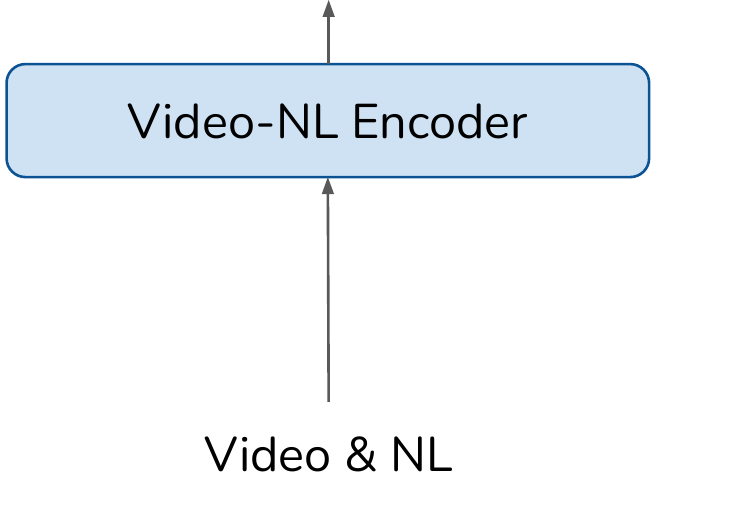}}
    \subfloat[]{\includegraphics[clip, trim=9cm 9cm 9cm 0cm, width=0.16\textwidth]{img/relworks.pdf}}
    \subfloat[]{\includegraphics[clip, trim=17cm 9cm 1cm 0cm, width=0.16\textwidth]{img/relworks.pdf}}
    
    \caption{Joint video-text embedding categories: (a) single-stream shared encoder; (b) separated-streams encoders with joint encoder; (c) separated-streams encoders with distance loss}
    \label{fig:embeddings}
\end{figure}

Inspired by the BERT model’s success for NLP tasks, numerous multimodal vision-language models have been proposed. Those kinds of methods, that try to learn a joint visual-textual embedding, can be divided into three categories (depicted in \Cref{fig:embeddings}): (i) single-stream shared encoder, (ii) separated-streams encoders with joint encoder; (iii) separated-streams encoders with distance loss. 

VideoBERT ~\cite{sun2019videobert} and ActBERT~\cite{zhu2020actbert} belong to the first category: they employ a single Transformer based encoder to learn joint embedding for video-text pairs, combining a sequence of ``visual words'', and a sequence of spoken words.
To create ``visual words'', Sun et al.~\cite{sun2019videobert} sub-sample the frames and apply a pre-trained video ConvNet, i.e. S3D~\cite{xie2018rethinking}.
Starting from the $BERT_{LARGE}$ model initialized with the pre-training, they added support for video tokens by appending 20,736 entries to the word embedding lookup table for each ``visual words''.
The novelty introduced by Zhu and Yang in ActBERT~\cite{zhu2020actbert} is the additional incorporation of global actions. 
They introduce a TaNgled Transformer block to encode features from three sources, i.e., actions embeddings, local regional objects obtained with Faster R-CNN network~\cite{ren2015faster}, and linguistic tokens obtained using Word-Piece~\cite{wu2016google}.

In the second category, three encoders are employed: one for the NL, one for the video, and a third one which receives the output of the previous two and learns a cross-modal embedding. The idea was introduced by Sun et al. with ``Contrastive Bidirectional Transformer'' (CBT)~\cite{sun2019learning}; other meaningful examples are HERO~\cite{li2020hero}, UniVL~\cite{luo2020univilm} and CLIPBERT~\cite{lei2021less}. HERO learns contextualized embeddings between the corresponding tokens and their associated visual frames. 
Instead, the peculiarity of CLIPBERT is that it randomly samples few short clips from the full-length videos at each training step. 

The latter category is the only one without a shared encoder: there are still two separate encoders for video and NL, and the resulting embeddings are correlated only through a loss function.  The idea comes from the Siamese network model, where the TML is applied to keep similar objects closer and move dissimilar ones further away. Unlike Siamese networks, the two branches are two separate architectures, each with its weights. In the method proposed by Miech et al.~\cite{miech2020end} the authors apply a 3D CNN backbone, specifically I3D~\cite{carreira2017quo}, for the video embedding and pre-trained word2vec~\cite{mikolov2013efficient} embeddings for NL. Then they introduce a specific loss, the Multiple Instance Learning Noise Contrastive Estimation Loss, to correlate the multi-modal information.
AYCE, the model proposed in this paper, belongs to this third category. 

\section{Data Analysis}
\label{sec:dataanalysis}

Before diving into the design of the architecture, we inspected the CityFlow-NL~\cite{feng2021cityflow} dataset. This section reports the insights that we have exploited to build the AYCE model. The training set is composed of $N=2498$ entries, and each entry $i$ has three objects: (i) three sentences $t_1^i, t_2^i, t_3^i$ in NL describing a Single Vehicle Tracking (SVT) sequence that appears in the video; (ii) a video of variable length; and (iii) the bounding box annotation of the tracked vehicle for each frame in the video. For all the information concerning (ii), (iii) and the video more in general, we will use the notation $v^i$ with $i \in \{1\,\dots,N\}$.
Each video sequence has in average $81$ frames (min: $1$, max: $3620$), while each sentence $t_j^i$ is composed in average of 9 words (min: $3$, max: $30$).

As also explained in the original work by Feng et al.~\cite{feng2021cityflow}, each sentence describes the target vehicle in terms of type of vehicle, color, and performed action. However, it also occurs that some of this information is missing at sentence level: the $0.76\%$ of the 7494 sentences misses the vehicle type, $4.68\%$ the vehicle color, and $1.53\%$ the performed action. However, given that a SVT sequence is described by a sentence-triplet and not only by a single sentence, it never occurs that those data are missing in all three sentences. Nevertheless, the three sentences are not consistent with each other. Inside a sentence-triplet, there are on average $2.07$ different target vehicle types, $1.85$ different colors, and $2.63$ different performed actions.
Therefore, it is clear that the sentence-triplet plays a central role in the model. 

Finally, we decided to analyze the heterogeneity of the data both at sentence-level and at sentence-triplet-level. We discovered that the sentences are very similar to each other. In particular, at sentence-level, the most recurrent sentence occurs $53$ times, while at sentence-triplet-level $15$ sentence-triplets have two identical sentences among the three, and in one case there are three identical sentences within a sentence-triplet. Then, we did the same test removing one of the previously mentioned features from the sentence, and we compared the remaining part. 
From this analysis, we discovered that there are up to $30$ sentence-triplets with two identical sentences (removing the type) which differ only by one, meaningful, word. The same behavior, even more obvious, can be noticed at sentence-level, where in some cases there are up to $172$ repetitions (removing the color).
\section{Method} 
\label{sec:method}

\begin{figure*}
    \centering
    \includegraphics[clip, trim=0.5cm 0cm 5.6cm 0cm, width=1.00\textwidth]{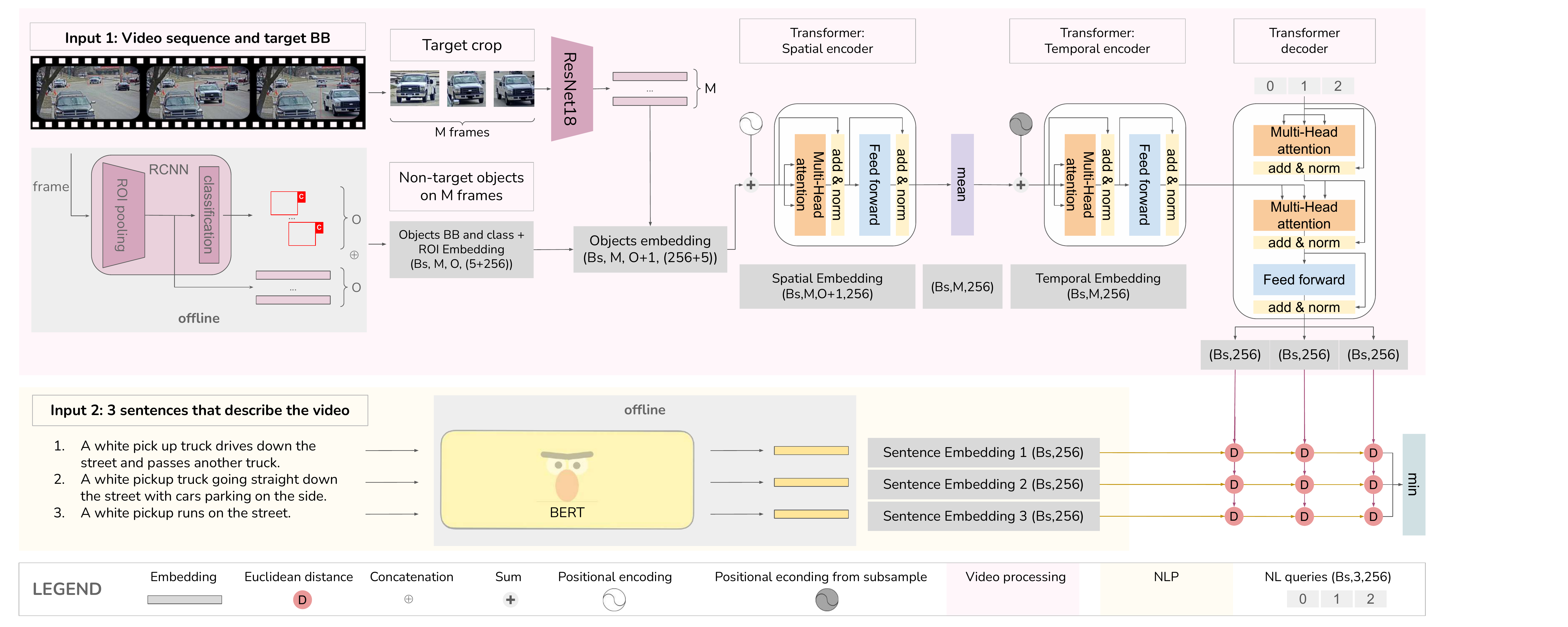}
    \caption{Overview of the best performing architecture with Visual Triple-Output and Language Triple-Output configurations.}
    \label{fig:model}
\end{figure*}
The proposed AYCE model is depicted in \Cref{fig:model}. 
It takes two separate inputs: (i) a \textbf{Visual} input, which includes the representation of a unique SVT sequence, and (ii) a \textbf{Natural Language} (NL) input, which includes the three sentences that describe the sequence. 
The final output is a distance value that quantifies the affinity degree between the two inputs: the lower distance is, the higher is the likelihood of the NL part to be a meaningful description for visual input. 
From an architectural standpoint, the Visual branch consists of a ResNet-family CNN, two stacked Transformers Encoders, and a final Transformer Decoder. Instead, the NL branch is a BERT model with the addition of a single fully connected layer. 
The two branches are trained on separate stages: first, the BERT model is fine-tuned on the NL descriptions; then, the Visual branch is trained from scratch using a metric-learning approach to learn an embedding of visual input projected onto the same latent space of the NL embeddings.

\subsection{Visual Branch} \label{sec:visual}
The $i$-th visual input $v^i$ is a tensor of shape $(M, O+1, (256+5))$ (omitting the batch dimension), where $M$ defines a subsample of the total number of frames in the \textit{i}-th vehicle tracking sequence, $O$ is the number of objects detected by an off-the-shelf object detector in the $a$-th frame and $(256+5)$ is the size of an object-embedding vector that encodes the representation of each $b$-th object, carrying visual (256), spatial (4) and classification (1) information. From now on, we will denote with $V^i$ the output of the visual-branch corresponding to $v^i$.\\

To keep a reasonable memory footprint, $M$ is capped to $80$, which is close to the average SVT length (\Cref{sec:dataanalysis}) hence sequences longer than 80 frames are online randomly subsampled with a uniform probability distribution.
The whole frame is never fed to the trainable model, but a set of \textbf{object embeddings} $O^a_b$ is collected for each of the $M$ selected frames: for each one of the $b$ objects detectable by a two-stage object detector on the \textit{a}-th frame, the corresponding representation is obtained by concatenating the detection bounding box with and additional visual-features vector, retrieved from the same detection model. The complete process is detailed in \Cref{subsec:objembs}. Conversely, the object embedding associated with the tracked vehicle is obtained by concatenating the provided tracking bounding box with a visual embedding vector computed by a dedicated convolutional backbone. This module, trained along with the rest of the model, takes as input a sequence of $M$ RGB frames depicting only the tracking vehicle, and produces a 256-dimensional visual representation.
Padding must be applied to $v^i$ both at batch level, since $M^i$ can differ between sequences, and at sequence level, given that different frames can spot a different number of detectable objects.

The visual input $v^i$ obtained in this way is fed through two identical Transformer encoders: the first \textbf{Spatial Encoder} operates on the sequence of $(O+1)$ objects, treating the remaining dimension as an additional batch shape. As in the original work~\cite{vaswani2017attention}, the output from the last block of the encoder has the same shape as the input, therefore the mean is computed over the object's axis to produce a fixed size embedding for each of the $M$ timesteps.  The averaged output of the Spatial Encoder results in a tensor with shape$(M, 256)$, which is fed to the \textbf{Temporal Encoder} that aggregates the temporal information of the visual input. Finally, a Transformer \textbf{Decoder} takes as input a \textit{NL-queries} set and the output of the Temporal Encoder to produce a tensor $(3,256)$ of three individual embeddings.

The rationale for this particular architecture is threefolds.   \textbf{(i)} Capturing fine-grained details about the \textit{appearance} of the tracked vehicle. By using a dedicated convolutional architecture to embed the sequence of the tracked vehicle crops we aim to properly distinguish very similar sequences which often differ only in details about the target vehicle (\textit{e.g.} the color or the type).   \textbf{(ii)} Modelling the \textit{spatial} relationships between the tracked vehicle and the surrounding objects. These sort of interactions are often highly discriminative of the described scene, given that a portion of the descriptions refers to interactions between the tracked vehicle and other road users ( \textit{e.g.} ``A $\langle$ \textit{target's description}$\rangle$ is [\underline{followed by} | \underline{following} | \underline{behind} | ...] a $\langle$ \textit{other vehicle's description} $\rangle$''). The addition of the precomputed object embeddings provides a strong insight regarding both the arrangement and the visual appearance of subjects involved in interactions with the target vehicle. %
   \textbf{ (iii) }Modeling the \textit{temporal} dimension. To truly capture all the spatio-temporal changes of the scene, the Temporal Encoder must operate on the output of the Spatial Encoder. In this way we expect the AYCE model to extrapolate trajectory details of the vehicles such as ``A $\langle$\textit{ target's description}$\rangle$ [\underline{turns left} | \underline{turns right} | \underline{going straight} | ...]''.

\subsubsection{Precomputed object embeddings} \label{subsec:objembs}

A pre-trained object detector is employed to obtain an additional set of detection bounding boxes and corresponding visual representations for each frame of the video sequences. We use the Faster R-CNN detector~\cite{ren2015faster} with a ResNet-50~\cite{he2016deep} backbone, trained on the 80-classes dataset COCO~\cite{lin2014microsoft}. Any detection with a confidence score lower than $0.85$ gets discarded. Among the remaining, we select only objects belonging to classes relevant to our problem, i.e. different types of vehicles, pedestrians, and other road objects ( e.g. \textit{stop sign} and \textit{traffic light}).

The two-stage nature of Faster R-CNN has been exploited to obtain an additional visual embedding for each detected object: right after the ROI-Pooling module, which takes the output of the Region-Proposal Network and outputs a $(7\times7\times256)$ vector for each proposed region, we apply an \textit{average pooling} operation to obtain an $(1\times1\times256)$ vector of features. 
Then, we concatenate the corresponding pooled feature vector to the values defining each detection bounding box. With this addition, the final output of the object detector becomes a $(5+256)$ vector with the first value being the object's class, the next four the box coordinates normalized by the image's shape in the range $(0,1)$ and finally the visual features vector.
We discard any detected box that might be associated with the tracked vehicle and, instead, we use the ground-truth tracking box provided by the dataset. However, since the class index is not provided we define a special index for the tracked vehicle. The visual features vector corresponding to the tracking box is initialized to zeros, to be then correctly filled by the convolutional backbone.

\subsubsection{Convolutional Backbone} \label{subsec:cnn}
The convolutional backbone is a plain ResNet-family convolutional network. Given the sequence of the $M$ sampled frames of a generic tracking sequence, the region delimited by the tracking bounding box is cropped from each frame, and the crops are resized to a common fixed size. The resulting tensor, of shape $(M, 3, W, H)$, is fed into the convolutional backbone. Each of the $M$ frames along the temporal axis is processed individually, treating the temporal dimensional as a batch dimension. The resulting tensor, of shape $ (M, 256)$, is used to correctly fill the target's vector feature previously initialized with zeros.

For the network, we chose the simplest approach of starting from a ResNet model pre-trained on Imagenet~\cite{deng2009imagenet}, in which we replaced the last fully connected layers used for 1000-classes classification with a newly initialized one, to provide an output of 256 elements.
Differently from the Faster R-CNN used to produce the object embeddings, the ResNet model is trained along with the remainder of the model.%

Considering the extreme range of variation in appearance due to orientation, lightning condition, occlusion, and several other factors, this component has been noted through qualitative observations to be the most influential to the overall effectiveness of the AYCE model.

\subsubsection{Stacked Spatio-Temporal Transformer} \label{subsec:encoders}

For the next part of our model, we use two identical Transformer Encoders in cascade. The first one aggregates the $(O+1)$ object embeddings for each of the \textit{M} frames into a single visual embedding, to provide a comprehensive frame-level representation. The second one aggregates the information along the temporal axis to infer the sequential meaning of the depicted scene. 

From a practical standpoint, we use the standard Transformer Encoder module~\cite{vaswani2017attention}. Therefore the number of Encoder blocks is set to $N_B=6$ and the number of Attention heads to $N_H = 8$ in all the Multi-Head Self-Attention (MHSA) modules. The architecture of the Transformer modules is kept unchanged, with the self-attention submodule and the feed-forward submodule, each followed by dropout with $p=0.1$, a ReLU activation function, and \textit{Add \& Layer Norm}. The bottleneck dimension in the linear submodule is also kept to the original value $2048$.

The first Encoder module, namely the \textbf{Spatial} Encoder, takes as input the complete visual input $v^i$, including the visual embedding produced by the CNN backbone.  Beforehand, a single linear layer is applied to reduce the embedding dimensionality from $261$ (or $5+256$) to $256$. This is required to have a size divisible by the number of MHSA Heads. Given the input tensor $(Bs, M, O', 256)$ where $Bs$ is the mini-batch size, the Spatial Encoder threats the number of frames (or timesteps) $M$ as an additional batch dimension. The information of the object embeddings is instead combined using the scaled dot-product Attention operations, carefully masking padded elements to avoid attending to padding values. Given that the Encoder's output is a tensor similar to the input, the \textit{mean} between the $O'$ embeddings is computed to produce a single embedding for each of the \textit{M} timesteps.

The subsequent \textbf{Temporal} Encoder takes as input the output of the previous one and operates almost in an identical way. The main difference is the definition of the \textit{Positional Encoding}: being \textit{M} a subset of a potentially significantly larger set (recalling that the longest sequence count 3620 frames, when $M$ is capped to 80), we use the indices of the sampled timesteps as input to the positional encoding, instead of the canonical sequence of incremental integers $(0, 1, ..., M-1)$. Using this \textit{sampling-aware} encoding we aim to make our AYCE model slightly more robust to information loss due to subsampling, as well as improving the overall generalization properties.

\paragraph{Visual Single-Output (VSO)}
We initially experimented with the simplest approach of producing a single embedding vector for the whole visual sequence (not depicted in \Cref{fig:model}), even though the NL branch might provide three different embeddings. The single embedding vector is obtained by applying a mean operator over the \textit{M} output of the Temporal Encoder's output.

\paragraph{Visual Triple-Output (VTO)}
Instead, to obtain three embedding vectors for the same visual input, we introduce an additional Transformer Decoder block which takes as input a set of \textit{NL-queries}. Similarly to the \textit{object-queries} proposed by Carion et al.~\cite{carion2020end}, \textit{NL-queries} are just a sequence of fixed length (equals to \textit{3}),  only consisting in the corresponding positional encoding.
The output of the Temporal Encoder is fed as \textit{encoder output} to the inner \textit{Encoder-Decoder Attention} modules of each decoder block. The output $V^i$ of the last decoder's block is a $(3,256)$-shaped embedding vector.\\
For the VTO approach, we use the notation $V^i_j$ to indicate the $j$-th row of $V^i$. Together with a careful redesign of the loss function, this second approach outperformed the former one.

\subsection{Natural Language Branch} \label{sec:nlb}
The NL branch takes care of encoding the NL descriptions and mapping them to the same latent space as the visual embeddings. This is necessary to compute a distance measure between the visual input and the textual description as output of the pipeline.
To fulfill this task we adopted the popular model BERT~\cite{devlin2018bert} as the core part of the AYCE model, with the addition of a single linear layer to reduce the dimensionality of the embedding from the original $768$ to the required $256$. Given the \textit{i}-th tracking sequence, we denote as $t^i_j$ the \textit{j}-th textual description out of the three associated ones, with $i \in \{1,\dots, N\}$ and $j \in \{1,2,3\}$. 

Similar to the distinction between VSO and VTO described for the visual branch we experimented with two different embedding representations.

\paragraph{Language Triple-Output (LTO)} This is the straightforward approach, each of the textual descriptions $t^i_j$ is independently processed by BERT producing three distinct $T^i_j $ embedding vectors.
\begin{equation}
    T_{j}^i = BERT(t_j^i) \quad \text{ for } j=1,2,3
    \label{BERT2}
\end{equation}

\paragraph{Language Single-Output (LSO)} In this alternative approach we experimented feeding a string-level concatenation to the language model, producing a single embedding for the whole tracking sequence.
\begin{equation}
    T_{1+2+3}^i = BERT(t_1^i \oplus t_2^i \oplus t_3^i)
    \label{BERT1}
\end{equation}
where $\oplus$ is the concatenation operation between strings.

From now on, since the largest portion of our experimentation is carried with the \textbf{LTO} approach, we refer to this solution unless specified.
Ideally we want the following property to be met:
\begin{small}
\begin{equation}
  \begin{cases}
    d(T^{i_1}_{j_1}, T^{i_1}_{j_2})  \ll  d(T^{i_1}_{j_1}, T^{i_2}_{j_2}) \\
    d(T^i_{j_1}, T^i_{j_2}) \approx 0
  \end{cases} \forall j: j_1 \neq j_2 \text{ and }i: i_1 \neq i_2
  \label{eq:bert_prop}
\end{equation}
\end{small}
In our experiments, the metrics used as distance are the Euclidean distance and/or the \textbf{Cosine Metric}~\cite{10.5555/576628}, an adaptation of the Cosine Similarity, in which a constant term equal to one is added to the opposite of Cosine similarity to obtain a measure whose resulting values are between $0$ and $2$.\\
Textual descriptions associated with the same tracking sequence should be mapped close to each other in the latent space, while the opposite must hold for descriptive sentences of distinct tracking sequences. We experimented with a popular implementation of BERT \cite{wolf-etal-2020-transformers} pre-trained on a corpus of English text with a Masked Language Model objective \cite{song2019mass}. As shown in \Cref{subsec:exp_bert}, the pre-trained model fails to achieve the property \Cref{eq:bert_prop}, hence in the next section, we detail the proposed solution to finetune the language model with a semi-supervised approach to make it suitable for our retrieval task.

\subsection{Optimization strategy}\label{sec:optimization}
\paragraph{Natural Language Branch.}
We decided to train the Visual branch and the NL branch in separate stages. For the latter, we aim to enforce the property defined in \Cref{eq:bert_prop}, via a BERT fine-tuning performed in a semi-supervised fashion.
To fulfill this purpose, we adopt the \textbf{Triplet Margin Loss} (TML) function\cite{chechik2010large,schroff2015facenet, schultz2004learning, weinberger2009distance}

\begin{scriptsize}
\begin{equation}
    \mathcal{TL}(A,P,N)=\frac{1}{Bs}\sum_{i=1}^{Bs}max(0,d(A^i,P^i)-d(A^i,N^i)+m)
    \label{eq:triplet_loss}
\end{equation}
\end{scriptsize}
By minimizing the TML, the distance from the baseline (anchor) input $A^i$ to the positive input $P^i$ is reduced to be less than the distance from the baseline (anchor) input to the negative input $N^i$ at least of $m$.

Given a tracking sequence, a random textual description $t^{i_1}_{j_1}$ is selected out of the tree to serve as anchor, another description $t^{i_1}_{j_2}$ is selected as positive example and a third description $t^{i_2}_j$ is sampled as negative from a randomly selected sequence. 

\noindent For the LSO approach, we use the same procedure. However, in this case, the positive input and the anchors are a concatenated random permutation of the same set of NL descriptions. Following the same concept, the negative input is the concatenation of the NL description set of a different sequence.
In \Cref{subsec:exp_bert} we detail the exact setup used for the final training of the visual branch, and we provide a comparison between the LSO and LTO approaches.

\paragraph{Visual Branch.}
The Visual branch is trained with a metric-learning objective: after defining a distance function between the embedding $V^{i_1}$ resulting from a visual input $v^{i_1}$ and a generic embedding $T^{i_2}$ obtained by the NL branch; the Visual branch is trained to reduce $d(V^{i_1},T^{i_2})$ with $i_1=i_2$ to be less than the same distance with $i_1 \neq i_2$ at least of $m$.

The exact definition of $d$ depends on the considered configuration of the Visual branch (VTO or VSO) and of the Language branch (LTO or LSO). A schematic summary of the explored combination is reported in \Cref{tab:model_opts}.

\begin{table}[!ht]
\centering
\small
\begin{tabular}{c|cc|cc}
\textit{Model}        & \multicolumn{2}{c|}{\textit{Visual Branch}}        & \multicolumn{2}{c}{\textit{Language Branch}}       \\
\multicolumn{1}{l|}{}              & \textbf{VSO  }            & \textbf{VTO}             & \textbf{LSO }              &\textbf{ LTO}              \\
\hline
VS-LT          &  \checkmark                &                   &                  & \checkmark                   \\
\hline
VS-LS         & \checkmark                 &                  & \checkmark                  &                     \\ 
\hline
\textbf{VT-LT}          &          & \checkmark                   &        & \checkmark                  \\              
\end{tabular}
\caption{Different experimented combination of Language and Visual branches.
}
  \label{tab:model_opts}
\end{table}

The best results have been obtained with the VT-LT approach, therefore hereafter the exact optimization process for this solution is detailed. 
The overall procedure is similar for other combinations, mainly differing for the definition of the distance measure used in the loss function.

The loss function used is the TML defined in \Cref{eq:triplet_loss}.  In this case, the anchor $A^i$ is defined as the Visual embedding $V^{i_1}$, the positive input $P^{i_1}$ is the Language embedding $T^{i_1}$ associated to the same sequence and the Negative input $N^{i_2}$  with $i_1 \neq i_2$ is the language embedding obtained from a different sequence.
Besides that, we sum a scaled Euclidean distance term that seeks to minimize the distance between the visual embeddings and the corresponding positive. This additional loss term is supposed to help to avoid local minima, given that during the optimization process only the anchor is moved while the language embeddings are fixed.

The comprehensive objective function $\mathcal{L}(A,P,N)$ is defined in \Cref{eq:loss}.
\begin{equation}
    \mathcal{L}(A,P,N)=\mathcal{TL}(A,P,N)+\frac{1}{Bs}\sum_{i=1}^{Bs}\beta \cdot \Phi(A^i,P^i)
    \label{eq:loss}
\end{equation}

The employed  distance function is an extension of Euclidean distance. Since in the VT-LT approach both the visual and the language branch provide three separate embedding vectors, the distance is computed for all the $9$ permutations, resulting in a matrix of distances $D^T_V$, in which each element is defined as:
\begin{equation}
D^T_V(m,n)=d(V^{i_1}_m ,T^{i_2}_n) ~~ m,n \in \{1,2,3\}.
\label{eq:dist_matrix}
\end{equation}
$\Phi(A^i,P^i)$ is defined as the minimum distance among the embeddings of the anchor and the positive NL  in $D^T_V$ , namely $min(D^{A^i}_{P^i})$. 
On the other hand, in $\mathcal{TL}$ we use the average distance among the embeddings of the anchor and the negative NL, namely $mean(D^{A^i}_{N^i})$. 
The value of $\beta$ is fixed to $\beta=0.1$.
In addition, we exploited a strategy of online hard-negatives mining as suggested by Xuan et al.~\cite{xuan2020improved}, selecting the negative value $N^i$ as $max(d(A^i, N^i))$ for $i$ in $(0, Bs-1)$.
In \Cref{subsec:model_experiments} we detail the evaluation process and present the results obtained on the CityFlow-NL dataset.

\section{Experiments and Evaluation} 
\label{sec:experiments}

\begin{table*}[th]
\label{tab:bert}
  \centering
  \small
  \begin{tabular}{c | ccccc | cc | cc}
  \textit{Model}
& \multicolumn{5}{c|}{\textit{Hyperparameters}}
& \multicolumn{2}{c|}{\textit{Intra-sentence-triplet}} 
& \multicolumn{2}{c}{\textit{Inter-sentence-triplet}} \\

 &\textbf{loss} &\textbf{margin} &\textbf{\# epoch} &\textbf{Bs} &\textbf{lr} & \textbf{mean}  & \textbf{var } & \textbf{mean } & \textbf{var } \\ \hline\hline
    $LTO_{BASE}$ & - & - & - & - & - & 0.1703 & 0.0068 & 0.1899 &0.0057\\
    $LSO_{BASE}$ & - & - & - & - & -& 0.0046 & 5.25E-06 &  0.1079 & 0.0015\\\hline
    $LTO^*_{FT}$ & Triplet &2.5 &4 &48 &1E-04 & \textbf{0.2089} & 0.0376 & \textbf{0.6140} &0.0723\\
    $LTO_{FT}$ & Triplet &2.5 &4 &48 &5E-05 & 0.2464 & 0.0444 &  0.6935 & 0.0838\\
    $LSO_{FT}$ & Triplet &2.5 &4 &20 &1E-04 & 0.0059 & 1.88E-05 & 0.4754 & 0.0409\\
    $LSO_{FT}^*$ & Triplet &2.5 &4 &20 &5E-05 & \textbf{0.0044} & 1.45E-05 & \textbf{0.4622} & 0.0260\\
  \end{tabular}
  \caption{BERT fine-tuning details. The first two rows report intra-sentence-triplet and inter-sentence-triplet distances for the BERT model without fine-tuning. In the rest, the hyperparameters and distances for the fine-tuned models can be found. The models marked with a $*$ are the ones used in the training of the Visual branch. }
  \label{tab:bertfinetuing}
\end{table*}
\begin{table*}[!ht]
    \centering
    \small
\begin{tabular}{c|c|c|c|c|c|c|c|c}
           & \multicolumn{5}{c|}{}                                    & \multicolumn{3}{c}{\textit{on Test set}}             \\
Name       & GPU      & Optimizer & ResNet   & crop size & initial LR & MRR             & Recall$@5$      & Recall$@10$      \\ 
\hline\hline
BASELINE \cite{feng2021cityflow} & -     & -      & - & - & -    & 0.0269 & 0.0264 & 0.0491  \\
VS-LT NO-FT & A100     & Adam      & Resnet18 & (80,80) & 1.0E-5    & 0.0224 & 0.0170 & 0.0358  \\
\hline
VT-LT-A-18 & A100     & Adam      & ResNet18 & (90, 110) & 3.5E-5     & \textbf{0.1078} & \textbf{0.1321} & \textbf{0.2491}  \\
VS-LT-A-18 & A100     & Adam      & ResNet18 & (90, 110) & 2.5E-5     & 0.0960          & 0.1189          & 0.2283           \\
VS-LT-A-34 & A100     & Adam      & ResNet34 & (90, 110) & 1.0E-5     & 0.0943          & 0.1283          & 0.2245           \\
VS-LT-S-18 & A100     & SGD       & ResNet18 & (90, 110) & 2.5E-4     & 0.0834          & 0.1189          & 0.2000           \\
VS-LS-A-34 & Titan RTX & Adam      & ResNet34 & (90, 110) & 7.5E-06    & 0.0738          & 0.1038          & 0.1679          
\end{tabular}
    \caption{Results obtained by the proposed architecture on the public leaderboard of the AI City Challenge Track 5. The first row reports the results obtained by the baseline model proposed by the dataset's authors, while the second reports the performance of the retrieval model without the BERT finetuning described in \Cref{sec:optimization}.}
    \label{tab:comparison}
\end{table*}

\subsection{BERT fine-tuning \label{subsec:exp_bert}}

To evaluate the fine-tuning of the NL branch we use the Cosine Metric to serve as a measure in Equation \ref{eq:bert_prop}. 
We refer to the average distance between descriptions associated with the same SVT as \textit{Intra-Tuple} or $d_{INTRA}$. On the other hand, \textit{Inter-Tuple} or $d_{INTER}$ is the average distance between descriptions of different sequences.
For both the LTO and LSO approach we compute mean and variance of $d_{INTRA}(T^i)$ for each \textit{i}-th input and $d_{INTER}(T^{i_1}, T^{i_2})$ for all possible combinations of ($T^{i_1}$, $T^{i_2}$), aiming to minimize the first and maximize the second. Based on those metrics we select a single best-performing model for each case to be then used in the training of the visual branch. \Cref{tab:bertfinetuing} reports a sample of the obtained results for different sets of hyperparameters.

\begin{figure}[!ht]
    \centering
    \subfloat[Without fine-tuning]{\includegraphics[clip, trim=1.2cm 1cm 0cm 0cm, width=0.25\textwidth]{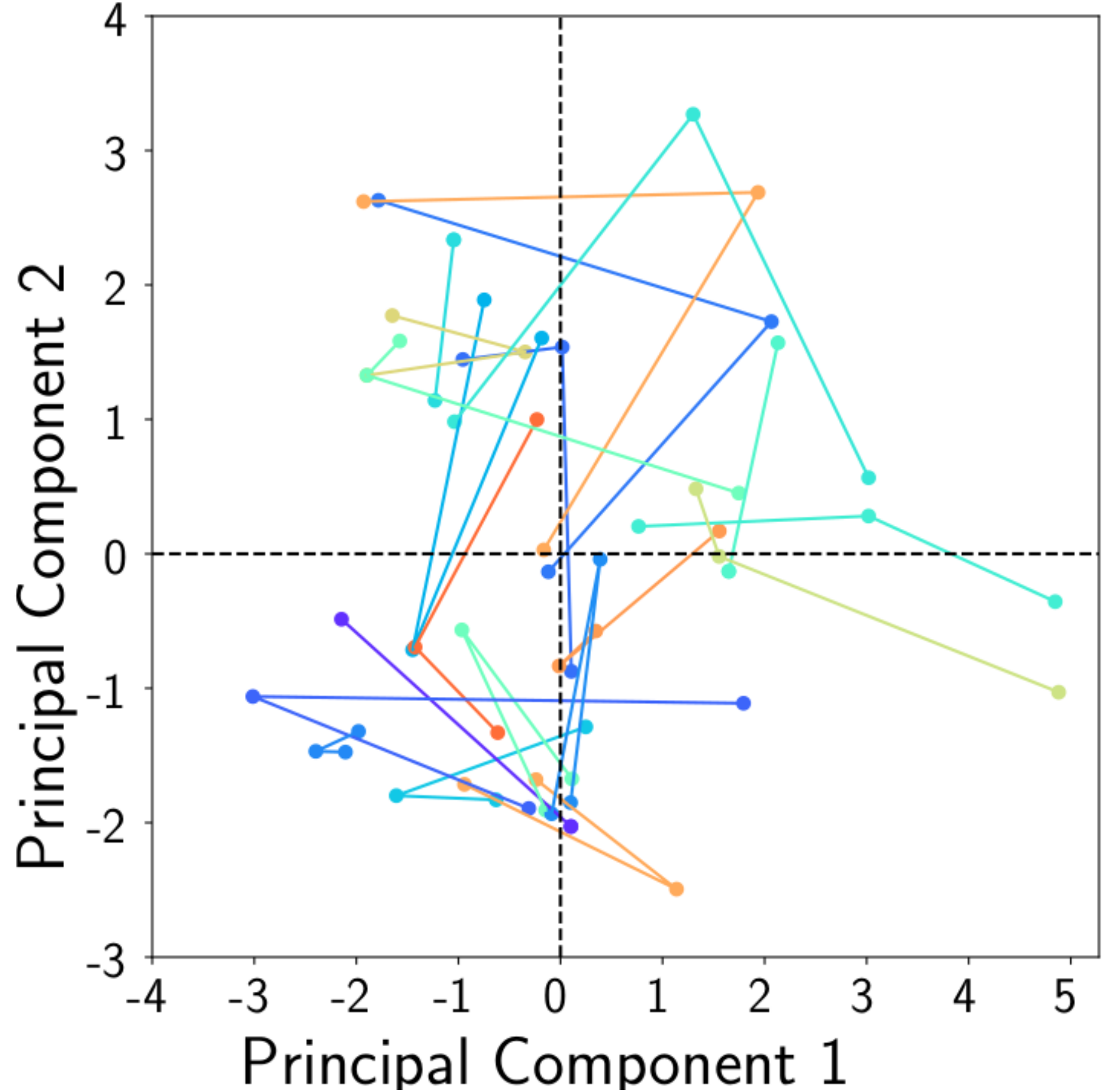}}
    \subfloat[With fine-tuning]{\includegraphics[clip, trim=1cm 1cm 0.2cm 0cm, width=0.25\textwidth]{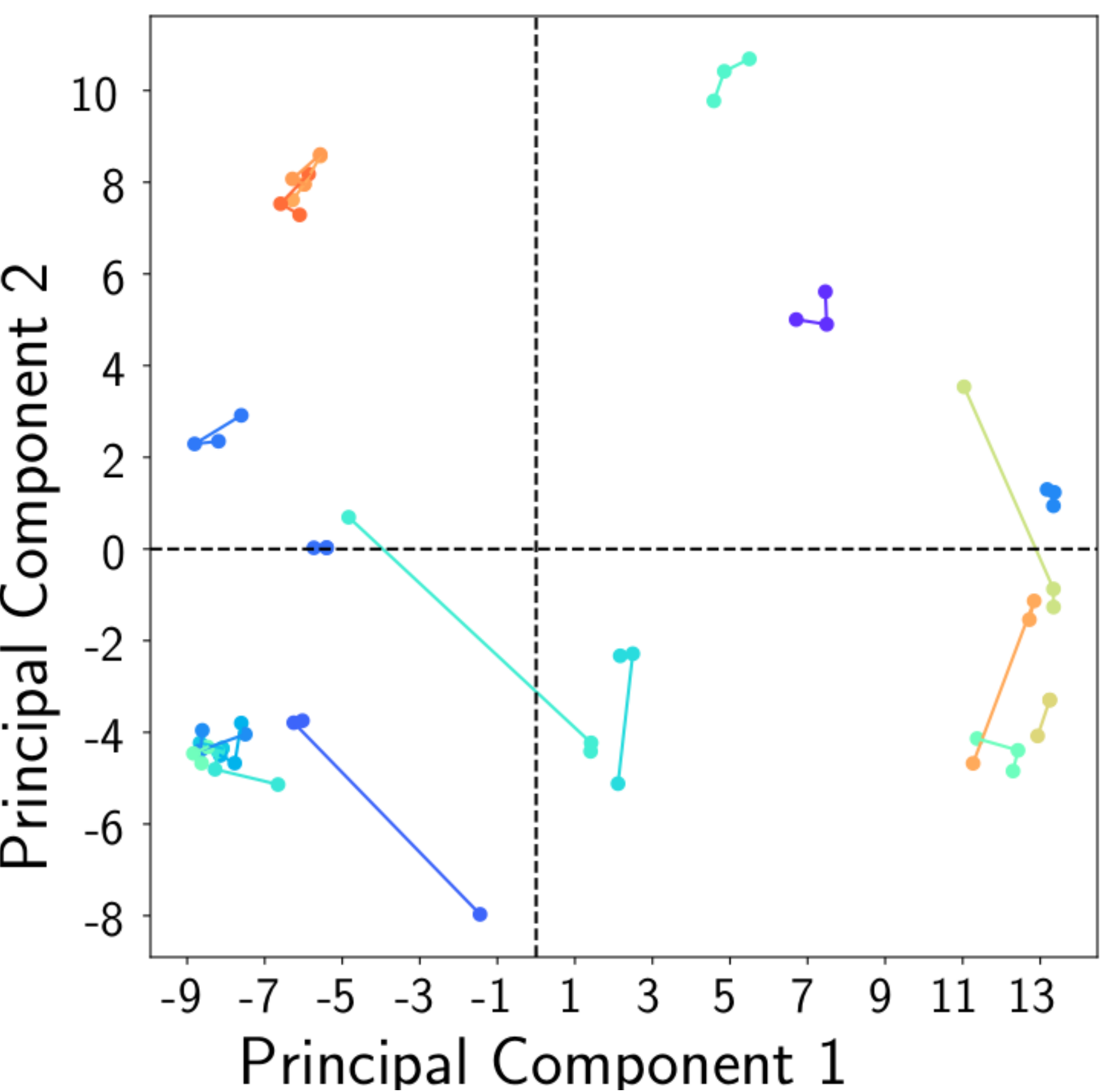}}
    
    \caption{2D representation of BERT embeddings, obtained via Principal Component Analysis.}
    \label{fig:bert_ft}
\end{figure}

The result of the fine-tuning is qualitatively represented in \Cref{fig:bert_ft}. 30 sentence-triplets have been projected into a 2D space before (a) and after (b) the $LTO_{FT}^*$ fine-tuning. After the fine-tuning the sentences of each triplet are close and far from the other ones; moreover, all the embeddings are better spread in the latent space. 

\subsection{Vehicle-Retrieval Model \label{subsec:model_experiments}}

To produce the submission result in the required format for AI City Challenge Track 5, we first
compute all the possible visual embeddings $V^i$ and language embeddings $T^i$ for $i$ in $(0, 530)$ ( where $530$ is the length of the test set). 
Since the two branches of the AYCE model are independent, the number of inference steps grows linearly in the number of sequences to be tested.
For each visual embedding $V^{i_1}$ we can compute the distance $d(V^{i_1}, T^{i_2})$ for each $i_2$ in the dataset.
At this point, for each  $V^{i}$ we sort the distance to the descriptions in descending order. In this way, it is possible to compute the Mean Reciprocal Rank (MRR) metric~\cite{feng2021cityflow}. In this case the distance function $d$ is defined as $min(D^V_T)$ with $D$ defined as in \Cref{eq:dist_matrix}.

We carried on preliminary experimentations, selecting a validation set by randomly sampling a $10\%$ of the provided training data, equal to $249$ tracking sequences. The best performing models were then trained from scratch on the full dataset. However, given the time constraints of the challenge, we could not pursue an exhaustive ablation study, and we hereafter report the explored experiments that led to the best results.
To avoid confusion, we report in \Cref{tab:comparison} only the results on the official test set, computed through the standard evaluation procedure of the AI City Challenge.
We compared the different architectural approaches summarized in \Cref{tab:model_opts}, on top of that we evaluated (i) the usage of ResNet18 rather than ResNet34 as CNN backbones (\Cref{subsec:cnn}), (ii) the usage of several optimizers, e.g. Adam, SGD with momentum and AdamW, (iii) the correct value for the margin value for the TML (\Cref{eq:triplet_loss}), (iv) other typical hyperparameters such as the learning rate $lr$. 

The results of the most successful methods on the test-set can be found in \Cref{tab:comparison}. The margin value $m$ is set to $1.0$ for all of them. It can be noticed that all the meaningful experiments outperform the baseline of the challenge~\cite{feng2021cityflow}.

The best model found follows the VT-LT configuration, using ResNet18 as convolutional backbone with input size equal to $(w, h)=(110, 90)$. It has been trained for a total of 680 epochs on two Nvidia A100 GPUs with a mini-batch size of 48 per GPU (totaling 96). Adam is used as optimizer and the learning rate has been initialized to $3.5e^{-5}$ and has been decreased to $2.5e^{-5}$ and $1.5e^{-5}$ at epochs $450$ and $650$ respectively. The total training time was around 27 hours.
\section{Conclusions} 
\label{sec:conclusions}
NL description offers a convenient and straightforward way to specify vehicle track queries. The AICityChallenge Track 5 aimed to perform vehicle retrieval given single-camera tracks and corresponding NL descriptions of the targets. After an exhaustive study of the state-of-the-art, we have developed AYCE: an original architecture, based on established and effective ideas, trained with an objective function suited for the considered task.%

Given the sheer size of the hyperparameters space, we consider the obtained results as preliminary and promising for future developments. However, we recognize the current result already quite adequate, since, with a $MRR > 0.1$, when querying with the NL descriptions the corresponding SVT is placed in the top 10 positions, on average. 

\section*{Acknowledgments}
This work has been partially supported by the CINECA grant number HP10BSTS2W and by POR-FSE 2014-2020 funds of Emilia-Romagna region (Deliberazione di Giunta Regionale n. 255- 30/03/2020).

{\small
\bibliographystyle{ieee_fullname}
\bibliography{bib}
}

\end{document}